%% file: neurips_2026.tex
\documentclass{article}

\usepackage[preprint]{neurips_2026}
\usepackage{booktabs, longtable, xcolor, array}
\definecolor{kiihl}{RGB}{210, 40, 60}
\usepackage[T1]{fontenc}
\usepackage[utf8]{inputenc}
\usepackage{enumitem}
\usepackage{CJKutf8}
\usepackage[most]{tcolorbox}
\usepackage{xcolor}
\usepackage{float}
\usepackage{wrapfig}
\definecolor{promptbg}{RGB}{245, 245, 245}
\definecolor{promptborder}{RGB}{200, 200, 200}
\usepackage{array}
\usepackage{amsmath}
\usepackage{amsfonts}
\usepackage{amssymb}
\usepackage{wrapfig}
\usepackage{mathtools} 
\usepackage[table]{xcolor}
\usepackage{url}
\usepackage{nicefrac}
\usepackage{verbatim}
\usepackage{subfigure}
\usepackage{anyfontsize}
\usepackage{bbding}
\usepackage{pifont}
\usepackage{wasysym}
\usepackage{microtype}
\usepackage{graphicx}
\usepackage{booktabs} 
\usepackage{multirow}
\usepackage{booktabs}
\usepackage{caption}
\usepackage{subcaption}
\usepackage{xcolor}
\usepackage{hyperref}
\usepackage{wrapfig}
\usepackage{xcolor}
\usepackage{enumitem}
\usepackage{fontawesome}
\usepackage{booktabs}
\usepackage{verbatim}

\definecolor{thaoblue}{HTML}{1D4ED8}

\definecolor{thaotodo}{RGB}{255,0,0}

\title{Personal AI Agent for Camera Roll VQA}

\makeatletter
\renewcommand\@fnsymbol[1]{}
\makeatother

\author{
  Thao Nguyen$^{1}$
  \hspace{0.6mm} Krishna Kumar Singh$^3$
  \hspace{0.6mm} Donghyun Kim$^2$
  \hspace{0.6mm} Yong Jae Lee$^{1, \dagger}$
  \hspace{0.6mm} Yuheng Li$^{3,\dagger}$\thanks{$\dagger$ \hspace{1mm} denotes equal advising; please contact \texttt{\{yuhli,krishsin\}@adobe.com} for dataset}
  \\
  \\
  $^1$University of Wisconsin-Madison
  \hspace{0.9mm}$^2$ Korea University
  \hspace{0.9mm}$^3$Adobe Research
  \\
  \\
    \url{https://thaoshibe.github.io/camroll}
}

\hypersetup{
    colorlinks=true,     
    linkcolor=magenta,   
    urlcolor=magenta     
}

\begin{document}

\maketitle

\begin{figure}[h]
    \centering
    \includegraphics[width=1\linewidth]{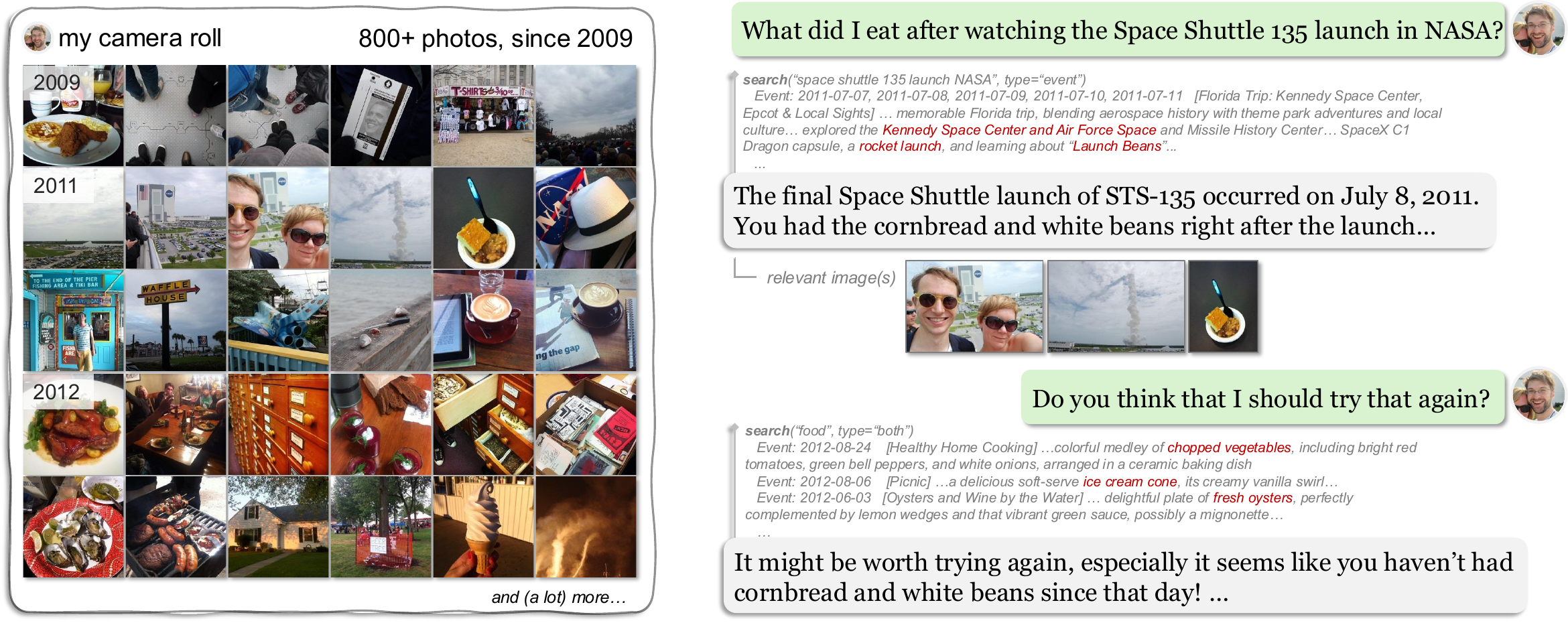}
    \caption{We study the VQA setting over the personal camera roll, where an AI assistant can search and retrieve relevant photos from thousands of user images, enabling more personalized responses.}
    \label{fig:teaser}
\end{figure}

\begin{abstract}
  We study the personal camera roll visual question answering setting.
  In this setting, a conversational AI assistant can access a user's personal camera roll and retrieve relevant photos to answer queries, ranging from simple factual questions (e.g., ``Name of the food I tried yesterday?'') to more open-ended ones (e.g., ``Recommend some dishes I have never eaten before'').
  Given the vast nature of the personal camera roll (i.e., multiple years, hundreds to thousands of photos), a successful AI assistant needs to understand a long-horizon, highly personalized visual content stream in order to navigate and locate the correct and/or relevant information.
  To support this, we collect and manually annotate questions that mimic real-world usage.
  The final dataset, \texttt{camroll}, contains 50 users, 31,476 images, and 2,500 QA pairs.
  We further design \texttt{camroll-agent}, a conversational AI agent equipped with hierarchical memory and a minimal set of tools for efficient navigation over large, personalized visual memory.
  Experimental results show that \texttt{camroll-agent} outperforms numerous baselines and methods for long-context understanding AI agents system.
  Together, the \texttt{camroll} dataset and \texttt{camroll-agent} highlight the gap in AI agents' long-context reasoning: personalized visual memory requires different approaches from standard long-context textual memory, especially when consistency, visual details, and user-specific context are present.

\end{abstract}

\input{sec/1_intro}
\input{sec/2_related}
\input{sec/3_data}

\input{sec/4_method}

\input{sec/5_experiement}
\input{sec/6_results}
\input{sec/7_conclusion}

\section*{Acknowledgment}
This work was supported in part by NSF IIS2404180, and Institute of Information \& communications Technology Planning\& Evaluation (IITP) grants funded by the Korea government (MSIT) (No. 2022-0-00871, Development of AI Autonomy and Knowledge Enhancement for AI Agent Collaboration and (No. RS-2025-2543949. Environment-Aware and Domain-Adaptive Multimodal Embodied AI for Real-World Interaction).

{
\small
\bibliographystyle{unsrt}
\bibliography{references}

}


\appendix

\input{sec/appendix}

\end{document}

%% file: sec/1_intro.tex
\section{Introduction}
\vspace{-2mm}

Take a moment to think about your camera roll.  
Chances are, it has become a growing digital archive of your life, filled with thousands of images: from the ordinary (e.g., yesterday's meal) to the significant, memorable events (e.g., your long-awaited visit to NASA).  
Multiple surveys report that smartphones, which have made taking photos easier than ever, enable users to actively take multiple photos daily, accumulating roughly 3,139 photos on each individual's phone~\cite{mixbook_survey}.  
These photos are not only external visual storage, but also powerful cues for autobiographical memory, enabling individuals to revisit past experiences~\cite{kislinger2021hunters,fernandez2024personal}.  
Yet, this promise is often undermined in practice.  
While 65\% of users share that they took photos in the first place to reflect later, more than half (55\%) feel overwhelmed when try to query about specific moments from their camera roll~\cite{photoaid_mobile_stats}.  
As a result, personal photo camera rolls increasingly resemble digital hoarding, despite arguably being the place with the richest and most densely informative records of one's life~\cite{affenstunde_digital_hoarding}.


But why is it so difficult to look back?  
First, this is because of the overwhelming volume: hundreds to thousands of photos, redundant or visually similar, scattered across multiple years.
Second, a typical camera roll nowadays (e.g., Google Photos, iPhoto) is mostly organized in chronological order and only supports basic similarity search (e.g., by people or places).
While helpful, this is not aligned with how humans naturally structure and recall memories (e.g., by context, experiences, goal-/ event-based).
Despite substantial progress in AI-powered tools for managing image collections (e.g., Apple Photos + Apple Intelligence~\cite{apple_intelligence}, Microsoft Copilot + OneDrive Photos~\cite{microsoft_copilot}), these systems still largely operate as a retrieval module at surface level (e.g., face/ object detection, or keyword-based search).
For example, one can search for ``NASA'' to find geo-tagged photos, but cannot ask more personalized and compositional questions such as: ``What did I eat after watching the Space Shuttle 135 launch?'', as this would require contextualizing the event and temporal order to retrieve the specific photo of the food (Fig.~\ref{fig:teaser}, right). Even further, to answer a follow-up question ``Do you think I should try that again?'' a model with knowledge that this user has not eaten the same meal since that day might respond differently, and less generically, than a model without such contextual awareness.
It would be a luxury to imagine a future where we can interact with an AI assistant (e.g., ChatGPT~\cite{chatgpt}, Gemini~\cite{gemini}, Claude~\cite{claude}) grounded in our personal camera roll.


From a technical perspective, one could naively feed all images into the context window of a MLLM.  
However, this quickly becomes impractical: a single HD photo costs 1-3k tokens, so a full camera roll of thousands images can easily reach \emph{1--10 millions tokens}!
This not only exceeds the context window of many models, but, even when feasible, significantly slows inference, and long-context understanding itself degrades as input length grows \cite{liu-etal-2024-lost,visual-in-haystack}.  
Alternative approaches leverage retrieval-augmented generation (RAG)~\cite{hipporag2,selfrag}, where the system builds a queryable database textual, and then retrieves a subset of relevant content (e.g., 1-3k tokens) at inference time.  
While efficient for long-text content, such designs can be misaligned with personal camera roll setting.
In particular, images are often treated as independent units, without incorporating personal context (e.g., events, relationships), which leads to noisy retrieval (e.g., querying ``my car'' returns all car instances regardless of ownership).
Moreover, majority of existing RAG-/ memory-based approaches~\cite{mem0,memos,simplemem,lightmem} only use generic image captions, discarding raw pixels, and therefore causing information loss.
For personal memory scenarios, fine-grained cues--such as identity, relationships, and event context--are often more important and relevant than explicit textual descriptions (e.g., ``me taking selfie with my partner'' vs. ``a selfie of a woman and a man'').

We argue that these limitations stem from the lack of appropriate data construction paradigms.  
There is currently no standardized framework for long-horizon personal visual memory.  
Existing datasets fall into three categories: (i) text-only personalization datasets~\cite{locomo,personamem,perltqa}, (ii) generic visual retrieval benchmarks without user-specific content~\cite{visual-in-haystack}, and (iii) real photo collections paired with simple retrieval queries~\cite{deepimagesearch,photobench}.
None of these captures the open-ended, personalized reasoning required for interacting with \emph{real camera rolls}. In practice, this direction has already begun to emerge in industry systems. For example, Google has introduced Gemini with Google Photos, enabling responses grounded in personal photo collections~\cite{google_ask_photos,google_gemini_nano_banana}; or Meta's Muse Spark supports connecting to personal albums or Facebook/ Instagram's posts~\cite{meta_muse_spark}. These efforts reflect growing interest in integrating MLLMs with personal visual data. However, little academic work studies how MLLMs reason over long-horizon personal visual streams, where information is fragmented across time and context. Bridging this gap is essential for developing a personalized AI assistant that can reliably and safely operate over real-world, long-horizon personal visual data.

In this paper, we take a step toward studying question answering over personal camera rolls.
We construct a dataset, \texttt{camroll}, from real user camera roll with annotated personalized visual question answering, and highlight the unique challenges that distinguish this setting from existing VLM benchmarks.  
Using \texttt{camroll}, we benchmark current systems on long-horizon understanding in personal visual image stream setting.  
We further design a conversational AI agent for this setting, \texttt{camroll-agent}, and analyze how it differs from conventional agents (e.g., coding agents).  
We argue that long-horizon, personalized understanding is a core capability of future personalized AI assistants, enabling more diverse and compelling applications (e.g., personalized consistent storytelling).

\vspace{-1mm}
In short, our contributions are as follows:
\vspace{-2mm}
\begin{itemize}[leftmargin=*, itemsep=1pt, parsep=0pt, topsep=2pt]
    \item We study personal camera roll VQA, requiring long-horizon and personalized visual reasoning.
    \item \texttt{camroll} dataset: 31,476 photos, 2500 QA pairs from 50 \emph{real} user camera roll.
    \item \texttt{camroll-agent}: conversational AI agent with: (i) hierarchical memory for efficient search/navigation; and (ii) a minimal set of tool to interact with large scale visual memory.
    \item Data insights and analysis, together with comprehensive benchmark results of existing methods, show the gaps in long-context personalized visual understanding.
\end{itemize}


\begin{figure}[t]
    \centering
    \begin{minipage}[t]{0.53\linewidth}
        \centering
        \includegraphics[width=\linewidth]{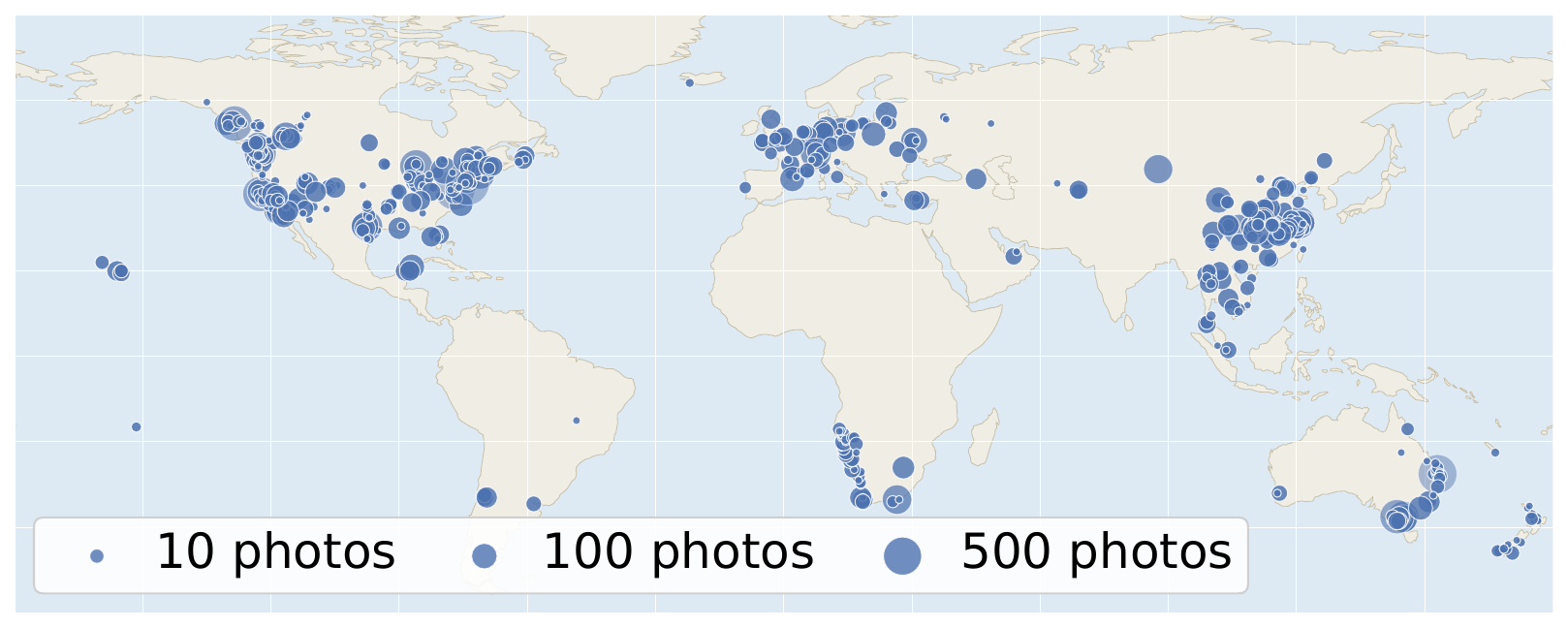}
        
        \vspace{-1mm}
    \end{minipage}
    \hfill
    \begin{minipage}[t]{0.46\linewidth}
        \centering
        \includegraphics[width=\linewidth]{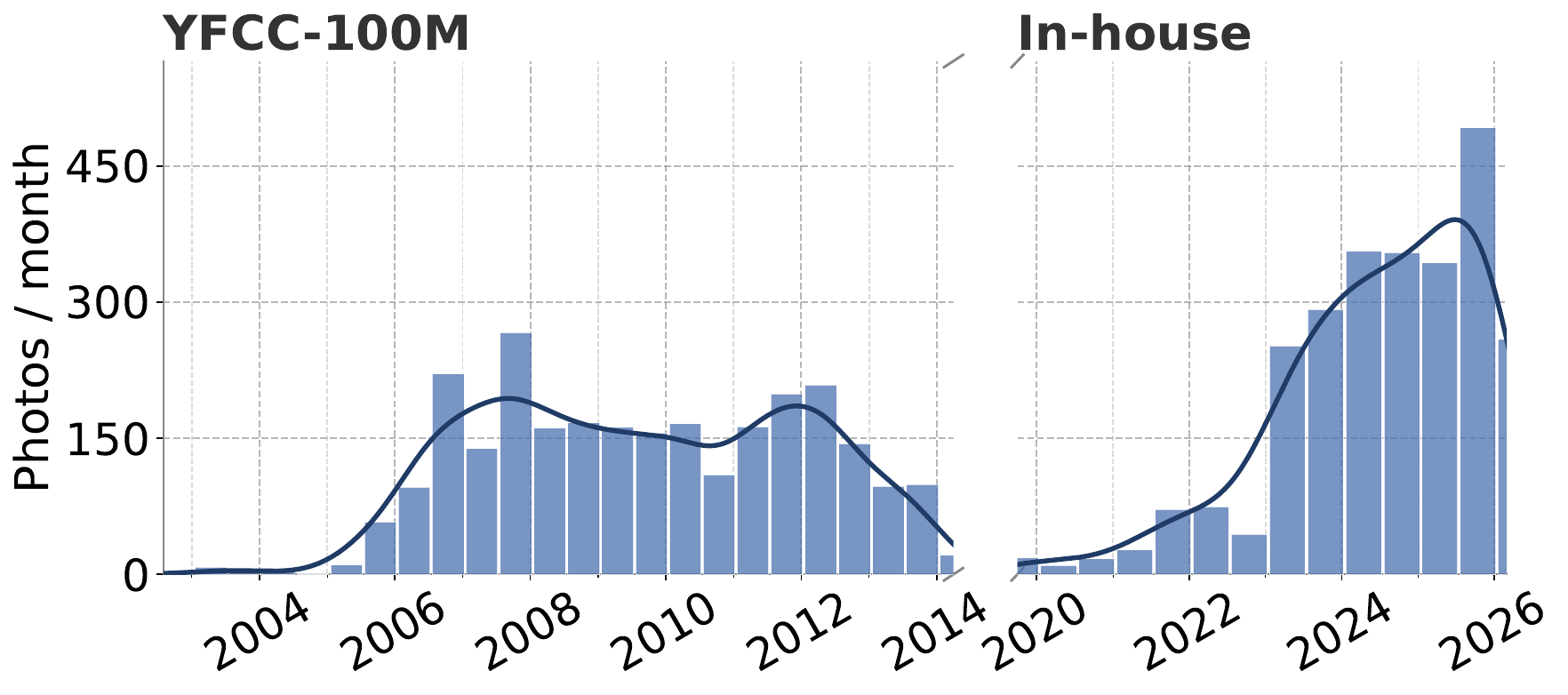}
        
        \vspace{-1mm}
    \end{minipage}

    \vspace{-2mm}
    \caption{Overview of \texttt{camroll}. Left: photos are captured across 25+ countries. Right: smartphone users (in-house subset) take substantially more images than digital camera users (YFCC-100M).}
    \label{fig:combined_demographic}
    \vspace{-5mm}
\end{figure}

%% file: sec/2_related.tex
\vspace{-2mm}
\section{Related Work}
\vspace{-2mm}

\textbf{Personal photo albums.}  Camera rolls, or personal photo albums, are extremely valuable digital assets of individuals, and have long been studied in computer vision. Early work focuses on relatively basic tasks, such as organizing photo collections, recognizing event types, and selecting representative or interesting images~\cite{Wang2017RecognizingAC}. 
Over time, the scope has expanded to more diverse settings, including leveraging related images within albums (or across a small set of albums) for image manipulation tasks such as inpainting or 3D generation~\cite{tang2023realfill}. 
More recently, with the rise of AI agents and large multimodal systems, there is increasing interest in working with personal images, enable the generic MLLMs to understand the personalized concepts~\cite{alaluf2024myvlm,nguyen2024yollava,yochameleon,PersonaVLM}. At the same time, growing attention has been devoted to long-context conversational reasoning and memory-intensive benchmarks~\cite{locomo,visual-in-haystack,personamem,nguyen2026personalvisualmemoryexplicit}. However, the majority of existing work focuses on internet-scale or conversational data, which often lack a coherent personalized visual stream (e.g., a daily random images, a road trip). 
There are also recent benchmarks for personal photo album retrieval~\cite{photobench,deepimagesearch}, though they primarily focus on retrieval rather than deeper understanding or reasoning over the collection.
In this paper, we pioneer the study of conversational VQA over personal camera rolls, a setting that requires understanding and reasoning across dense personal visual narratives.

\textbf{MLLMs with long-context understanding.}
With the rapid development of multimodal large language models (MLLMs), there has been continuous progress in understanding long-context inputs, including pure text, interleaved multimodal sequences, and image collections. 
A consistent observation is that model performance degrades as the context length increases~\cite{liu-etal-2024-lost,Du2025}. Alongside efforts to extend context windows and improve efficiency, retrieval-augmented methods—more broadly framed as memory mechanisms—have emerged as a promising solution to mitigate these limitations~\cite{selfrag,memorybank,hipporag2}. 
While such approaches achieve strong performance on text-centric benchmarks (e.g., LOCOMO~\cite{locomo}), they are less effective for images. This is largely because images are typically converted into textual captions and then processed as text. 
In contrast, we treat images as a first-class modality, indexing and reasoning over them directly rather than reducing them to text.


\textbf{AI Agents.} 
AI agents extend passive LLMs into autonomous systems capable of reasoning, planning, and executing multi-step actions to achieve goals~\cite{ReAct}. A typical agent consists of: (i) LMM/MLLM as the core reasoning engine; (ii) tools that enable interaction with external environments (e.g., file systems); and (iii) memory, which maintains context across interactions for long-term consistency and personalization. 
Recent progress has been particularly strong in domain-specific agents, such as coding agents (e.g., ClaudeCode~\cite{anthropic2025claudecode}), which operate in well-defined environments. While these systems can sometimes generalize to other tasks (e.g., travel planning), in practice, different domains require substantially different tools and interaction patterns. As a result, truly general-purpose agents remain limited. 
In most current systems, tools are manually designed and iteratively refined through trial-and-error, often guided by failure cases. 
In line with recent efforts toward more personalized and task-oriented agents, we explore the design of an AI agent tailored for personal camera roll.

%% file: sec/3_data.tex
        
        


\vspace{-2mm}
\section{Camroll: Personal Camera Roll Dataset}
\vspace{-2mm}

\texttt{camroll} is a personal camera roll question answering dataset.
Each camera roll contains photos naturally taken by a user via personal devices (e.g., mobile phones), over 2-6 years, paired with corresponding annotated  QA pairs.
At the time of writing, \texttt{camroll} comprises 50 users, 31,476 images, and 2,500 QA pairs drawn from two sources (in-house and YFCC-curated).



\vspace{-2mm}
\subsection{Data collection and annotation}
\vspace{-2mm}

\textbf{Source.} \texttt{camroll} is derived from two sources: (i) the publicly available YFCC-100M~\cite{yfcc100m}; and (ii) purchased from real users. While YFCC provides large-scale public multimedia collection, it is significantly outdated (up to 2014) and biased toward professional photography, making it less aligned with average personal camera roll. By comparison, the in-house data better reflects current in-the-wild mobile capture patterns, which are more incidental, redundant, and less curated.

\textbf{Filtering.} To construct natural personal camera rolls that reflect users' daily lives, we apply three strict criteria: (i) more than 500 photos per user; (ii) a temporal span of at least 2 years; and (iii) all images released under Creative Commons licenses, thus suitable for research use.
Since YFCC-100M is dominated by themed and professional photography, we further apply a multi-stage filtering pipeline to surface camera-roll-like users. This pipeline combines metadata-level constraints (e.g., upload volume, activate days, etc) with an LLM-ensemble judgment that retains only users whose photo collections exhibit rich personal-life traces.
We then randomly sample 20 users meeting above criteria and download all their images, yielding 15,927 images (see Tab.~\ref{tab:yfcc_license_distribution} for full license distributions).
For in-house data collection, we recruit participants under the same criteria and request access to their mobile camera rolls, along with permission to use the data for research purposes. Participants may review and remove any images prior to submission. In total, 30 participants contribute 15,658 images.
The final \texttt{camroll} dataset consists of 50 personal camera rolls, each paired with a profile photo representing its owner. Every image is timestamped in \texttt{YYYY-MM-DD HH:MM:SS} format.

\textbf{Annotation Protocol.} Collecting meaningful and personalized questions over long-term personal photo camera rolls is challenging.
Ideally, the most faithful questions would be posed by the photo owners themselves, as they uniquely understand the context, intent, and circumstances behind each capture. However, such annotations are not scalable (e.g., ATM-Bench~\cite{atm} is first-author annotating his own data). An common alternative is to synthesize queries using LLM-based pipelines~\cite{deepimagesearch,photobench}. While effective for visually grounded retrieval (e.g., ``photos with silver heart-shaped bracelet''~\cite{photobench}), these methods often fail to capture higher-level or longitudinal questions (e.g., ``Am I losing weight in recent years?'').
Instead, we prioritize human-posed questions. Annotators review full personal photo collections and are instructed to imagine living the subject's life, then generate natural questions they would ask an AI assistant. To ensure quality and consistency, the annotation process includes multiple rounds of guideline calibration, with feedback incorporated between rounds.

\textbf{Questions.} Humans organize memory into two primary systems: semantic memory, which captures general knowledge and abstract facts, and episodic memory, which encodes specific events situated in time and place~\cite{tulving2002episodic}. Motivated by this distinction, and by the nature of personal camera rolls which can reflect both personal identity and life trace, we collect two corresponding types of questions: (i) semantic and (ii) episodic. Annotators generate questions in two categories: (i) \emph{semantic} questions about the person that are not tied to a specific event or moment, and (ii) \emph{episodic} questions grounded in specific past events.  For each camera roll, annotators produce 10 semantic and 40 episodic questions. Episodic questions must be explicitly supported by a set of images (i.e., evidence), indicating how the answer can be inferred. This design ensures that all questions are human-realistic and factually grounded, which is critical for evaluating AI agents that aim to reason over personal visual histories.


\textbf{Answers.}
Following the \cite{vqa} protocol, annotators are asked to provide a concise, factually correct answer (i.e., a \emph{golden answer}) in the form of a short phrase. We additionally ask annotators to create two incorrect answers to construct a 3-option multiple-choice format.
When applicable, annotators also select the images they used to form the question and answer, referred to as \emph{gold evidence(s)}.






\vspace{-2mm}
\begin{table}[ht]
\centering
\small

\begin{minipage}{0.45\textwidth}
\centering
\captionof{table}{Embedding-level personalization measured by kNN user purity. Questions exhibit substantially stronger user-specific patterns than answers.}
\setlength{\tabcolsep}{2pt}
\resizebox{\linewidth}{!}{
\begin{tabular}{lrrrr}
\toprule
\textbf{Subset} & \textbf{N} & \textbf{Baseline} & \textbf{Question} & \textbf{Answer} \\
\midrule
All dataset          & 2500 & 1.96\% & 13.74\% & 4.13\% \\
\hspace{1.5mm}\textit{Semantic} & 500  & 1.80\% & 2.08\%  & 1.94\% \\
\hspace{1.5mm}\textit{Episodic} & 2000 & 1.95\% & 16.46\% & 4.26\% \\
\bottomrule
\end{tabular}
}
\vspace{1mm}

\label{tab:knn_purity}
\end{minipage}
\hfill
\begin{minipage}{0.53\textwidth}
\centering
\captionof{table}{Fractional-$k$ answer coverage across datasets. \texttt{camroll} exhibits substantially higher answer diversity compared with existing VQA datasets.}
\setlength{\tabcolsep}{2pt}
\resizebox{\linewidth}{!}{
\begin{tabular}{lcc>{\columncolor{gray!7}}r}
\toprule
\textbf{Diversity metric $\downarrow$} & \textbf{VQA}~\cite{vqa} & \textbf{LLaVA}~\cite{llava} & \textbf{\texttt{camroll}} \\
\midrule
Top-0.1\% coverage & 56.22 & 51.30 & 2.96 \\
Top-0.5\% coverage & 68.22 & 54.92 & 8.44 \\
Top-1.0\% coverage   & 73.52 & 56.34 & 12.84 \\
Top-5.0\% coverage   & 85.71 & 62.09 & 24.52 \\
Top-10.0\% coverage  & 89.85 & 65.87 & 32.04 \\
\bottomrule
\end{tabular}
}
\label{tab:fractional_coverage}
\end{minipage}

\label{tab:data_personalization}
\vspace{-2mm}
\end{table}

\subsection{Personalization characteristics in \texttt{camroll} dataset}

Beyond the general statistics discussed in Appendix~\ref{appendix:sec:dataset}, we further analyze the personalization characteristics in \texttt{camroll} by examining whether questions and answers exhibit user-specific patterns.


\textbf{Embedding-level personalization.}
Each question and answer is embedded by BGE-M3~\cite{chen-etal-2024-m3}. We compute kNN user purity at $K{=}10$, defined as the fraction of nearest $K$ neighbors belonging to the same user (random baseline:  $\sim$2\% for 50 users). As shown in Tab.~\ref{tab:knn_purity}, episodic questions reach \emph{16.5\%} purity ($8{\times}$ above baseline), indicating strong user-specific patterns. In contrast, semantic questions remain near baseline (2.1\%), as they capture general persona questions shared across users (e.g., hobbies). Answer purity is lower in both cases (4.3\% episodic, 1.9\% semantic). The asymmetry is structural: questions typically carry a layer of user-specific contextual signals---including recurring proper nouns, event anchors---which causes embeddings from the same user to cluster naturally. In contrast, answers are often bare values (e.g., ``Tokyo'') and thus disperse by topic rather than by user.

\textbf{Value-level personalization.}
At the level of discrete answer strings, \texttt{camroll} exhibits strong user-level disjointness. Of the $1{,}875$ distinct gold answer strings, $90.2\%$ appear in only one user's roll. The same pattern holds at finer granularities: $66.9\%$ of distinct content tokens (length $\geq 4$) and $88.1\%$ of unique answer bigrams are tied to a single user. This provides a complementary perspective to embedding-based analysis: semantically similar answers (e.g., ``Stanford'' and ``Tsinghua'') may lie close in embedding space, yet remain entirely user-specific in occurrence. 

\textbf{Cross-dataset comparison.}
To provide a comprehensive view of \texttt{camroll}'s answer distribution, we compare it with VQA~\cite{vqa} and LLaVA-1.5-mix-665k~\cite{liu2023improvedllava}, focusing on long-tail behavior. We report \emph{fractional-$k$ coverage}: the fraction of all answer occurrences captured by the top-$x\%$ most frequent answers in each dataset's vocabulary. 
The contrast is sharp (Tab.~\ref{tab:fractional_coverage}): the top $10\%$ of the vocabulary covers 89.9\% of answer tokens in VQA and 65.9\% in LLaVA, but only \emph{32.0\%} in \texttt{camroll}. At the head, the gap is even more pronounced---the top 0.1\% answers account for over half of all answer occurrences in VQA and LLaVA, but only \emph{2.96\%} in \texttt{camroll}. This heavy-tailed distribution arises from user-specific value supports: each user's camera roll induces its own localized answer distribution, resulting in a globally diverse but individually concentrated vocabulary!

%% file: sec/4_method.tex
\section{Camroll-agent: A Personal Camera Roll Agent}
\label{sec:approach}

We introduce \texttt{camroll-agent}, a conversational agent that answers
questions over a user's personal camera roll
$\mathcal{I}=\{I_i\}_{i=1}^N$. The agent is built on two ideas. First, we
construct a \emph{hierarchical personal memory} that lifts raw pixels into
two progressively more abstract layers
(Sec.~\ref{sec:approach:memory}).
Second, we expose this memory through a set of dedicated tools
organised along a principled two-axis design
(Sec.~\ref{sec:approach:tools}).

\vspace{-2mm}
\subsection{Hierarchical Personal Memory}
\label{sec:approach:memory}

\begin{figure}[t]
    \centering
    \includegraphics[width=1\linewidth]{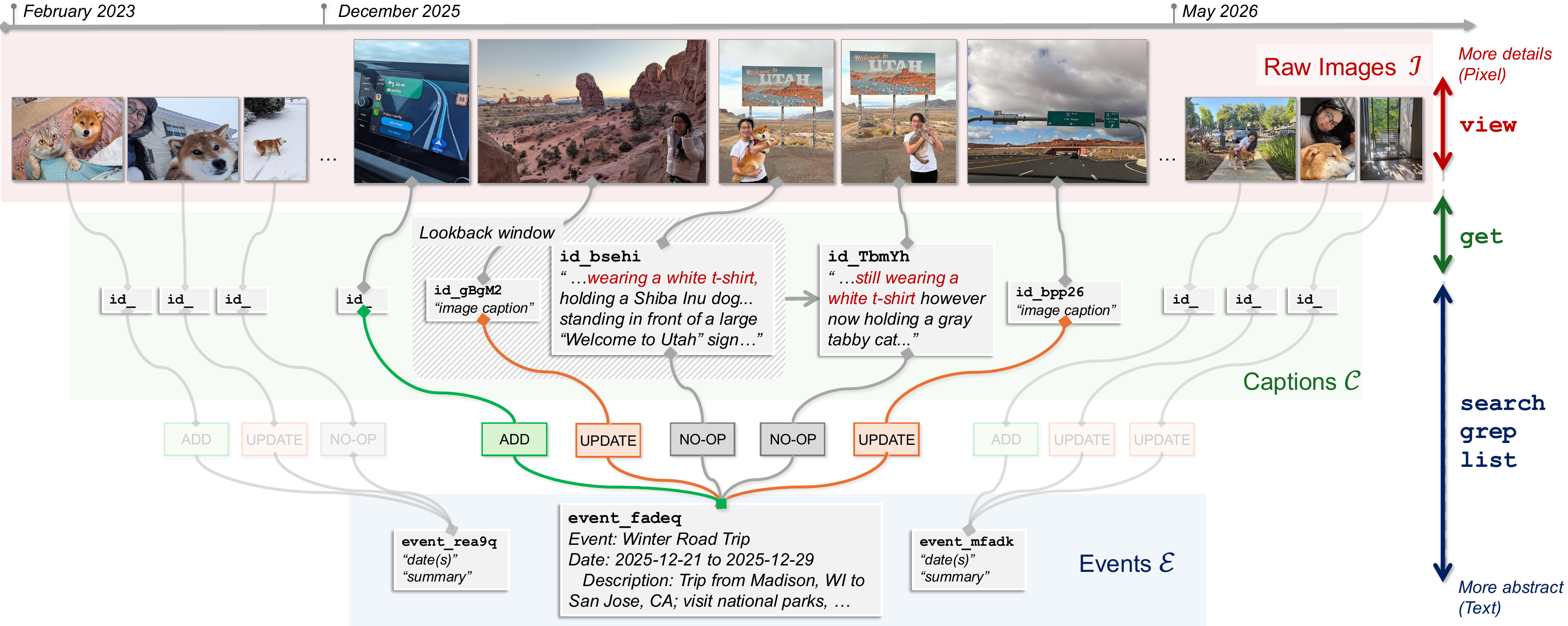}
    \caption{Hierarchical memory for personal camera rolls, organized from low-level visual pixels ($\mathcal{I}$) to higher semantic abstractions (captions $\mathcal{C}$, events $\mathcal{E}$). Agent interactions are designed accordingly, ranging from expensive tool (\texttt{view}, \texttt{get}) to cheaper one (\texttt{search}, \texttt{grep}, \texttt{list}).
    }
    \label{fig:approach}
    \vspace{-7mm}
\end{figure}

\textbf{Three-level pyramid.}
We organise memory as a three-level pyramid that abstracts upward from
concrete pixels to compact episodic units, while preserving full
links between adjacent levels (Fig.~\ref{fig:approach}):

\vspace{-2mm}
\begin{itemize}[leftmargin=*, itemsep=1pt, parsep=0pt, topsep=2pt]
    \item \textbf{Pixels} $\mathcal{I}=\{I_i\}_{i=1}^N$: raw photos kept
    untouched on storage.
    \item \textbf{Image captions} $\mathcal{C}=\{c_i\}_{i=1}^N$: personalized caption and per image metadata
    (timestamp, location).
    \item \textbf{Event summaries}
    $\mathcal{E}=\{e_j\}_{j=1}^M$, where each
    $e_j=(\mathcal{I}_j,\,d_j,\,m_j)$ groups a chronologically
    contiguous subset $\mathcal{I}_j\subseteq\mathcal{I}$ with a
    natural-language summary $d_j$ and metadata $m_j$
    (date, location).
\end{itemize}
\vspace{-2mm}


We construct abstract layers by processing the camera roll in chronological order, as described below.

\textbf{Personalized captions.}
Generic captions describe a photo from no one's point of view. To make
them useful as personal-memory cues, we condition the captioner on the
user's identity and recent visual context. For each image $I_t$ we feed
the captioning MLLM: (i) the user's profile photoand (ii) a
\emph{look-back window} of the most recent $k$ images $\{I_{t-i}\}_{i=1}^k$.
This grounds the caption in who the photo is of (the user vs.\ a
stranger) and what was happening just before, reduces the relevant noisy details.


\textbf{Event segmentation}. 
To form events, we prompt an MLLM to process images in an incremental fashion, with the goal of detecting episodic memory units (e.g., a trip, a wedding). Given the current image caption $c_i$, its timestamp, the most recent $k$ image captions $\{c_{i-j}\}_{j=1}^k$, and the summary of the current (most recent) event $e_m$ where $m = |\mathcal{E}|$, the MLLM chooses one of the following actions:

\begin{description}[itemsep=2.5pt, topsep=1pt, parsep=0pt, partopsep=0pt]
  \item[\texttt{ADD}.] Create new event $e_{m+1}=(\{I_i\},d_{m+1},m_{m+1})$
  when $I_i$ starts a new episode (e.g., new trip).
  \item[\texttt{UPDATE}.] Extend the current event,
  $\mathcal{I}_m\!\leftarrow\!\mathcal{I}_m\cup\{I_i\}$, and rewrite
  $d_m$ when $I_i$ refines or extends the same broader episode (e.g.,\
  a new day of a multi-day trip).
  \item[\texttt{NO\_OP}.] Append $I_i$ to the current event without
  rewriting $d_m$, when $I_i$ adds nothing new to the summary
  (e.g.,\ another selfie at the same place).
\end{description}
The first image is forced to \texttt{ADD} since $\mathcal{E}=\emptyset$.
The exact prompt is given in Appendix~\ref{appendix:sec:prompt}.

\textbf{Cross-linked storage.}
Every record receives a stable hashed ID
(\texttt{id\_<h>}, \texttt{ev\_<h>}). Each image stores the
\texttt{event\_id} of its parent event, which gives O(1) bidirectional
navigation: from any image we can look up its event in one hop, and
the set of images of an event is recovered by reverse lookup on the
foreign key. This invariant lets the agent move freely across the
pyramid without bespoke joins.
It is worth to mention, while mainly described here for personal camera rolls with images only, the same design extends naturally to other personal data modalities (i.e., emails).

\vspace{-2mm}
\subsection{Designing Tools for Memory Access}
\label{sec:approach:tools}





While the hierarchical memory organizes personal camera roll into structured representations, the \texttt{camroll-agent} still requires an efficient and budget-friendly interface to access it.
This motivates a small set of tools, which we design by decomposing the design space along two orthogonal axes: (i) \emph{retrieval paradigm}--how candidate records are retrieved (semantic, lexical, or filtering); and (ii) \emph{access depth}--at which granularity level (preview, full text record, or raw pixels).

\textbf{Tools.} This factorisation yields five tools. The matchers \texttt{search}, \texttt{grep}, and \texttt{list} cover complementary retrieval paradigms and return lightweight text previews. \texttt{get} upgrades a result to its full text record, while \texttt{view} enables direct inspection of raw images.

\begin{description}[
    labelwidth=5mm,
    leftmargin=6mm,
    labelsep=1mm,
    itemsep=2.5pt,
    topsep=1pt,
    parsep=0pt,
    partopsep=0pt
]
  \item[\texttt{search(query)}.] \hspace{1mm}For semantic search, all records are enriched with metadata and embedded using a frozen text encoder. Query is encoded with the same model, and the top-$k$ most similar records are retrieved via cosine similarity, each shown with a short preview (e.g., truncated captions).

  \item[\texttt{grep(keyword)}.] \hspace{1mm}For lexical search, exact or verbatim queries (e.g., ``NeurIPS''), semantic similarity is unreliable. In cases requiring exact token matching, \texttt{grep} performs BM25 retrieval to return the top-$k$ lexically matching records, each also with a short preview (e.g., truncated captions).

  \item[\texttt{list(condition)}.] \hspace{1mm}For structured filtering, many memory questions naturally impose metadata constraints such as time and location (e.g., ``in late October 2021'', ``in Paris'') rather than referring to content. \texttt{list} applies simple metadata filters to retrieve matching records.

  \item[\texttt{get(id)}.] \hspace{1mm} For full-text rendering, as the above tools only return short previews. \texttt{get} takes an \texttt{id} to fetch the full stored text (e.g., full caption, image paths). This preview/full split keeps each exploration withinin token budget, while still allowing agent to ``zoom in'' record of interest.

  \item[\texttt{view(id, prompt)}.] \hspace{1mm} For raw pixel-level inspection. Some questions require visual details that captions do not preserve (e.g., fine-grained, OCR). \texttt{view} re-examines the original images at query time: it takes a list of \texttt{id}s (up to six per call) together with a question prompt, and returns a VLM-generated textual analysis. Since image understanding is substantially more expensive than text retrieval, \texttt{view} is used only when textual evidence is insufficient.
\end{description}


\textbf{Interaction protocol.}
\texttt{camroll-agent} follows a standard ReAct~\cite{ReAct} loop. The agent is initialized with a system prompt, a description of the memory schema, and tool descriptions. At each step, agent produces a thought, and then either issues a tool call or emits a final answer. We additionally append a budget reminder (``step $T$, tool budget: $x/y$ remaining'') to encourage efficient tool use. Tool outputs are returned in a uniform format and appended to the interaction history. Tool outputs are returned in a uniform format and appended to the interaction history. The loop terminates either upon a final answer or when the step budget is exhausted, in which agent must answer without further tool use.

\textbf{Compatibility.}
All interaction is mediated through these five tools, so the design of \texttt{camroll-agent} is model-agnostic: swapping the LLM, the captioner, or the retrieval backend requires only replacing the corresponding component, while the agent loop and the tool interface stay unchanged. This modularity enables the cross-system comparisons in Sec.~\ref{sec:ablation}. We expect that jointly fine-tuning the LLM and the tools would yield further gains and leave this for future work.


%% file: sec/5_experiement.tex
\vspace{-2mm}
\section{Experiments}
\vspace{-2mm}
\subsection{Experimental Settings}
\label{sec:experimental_settings}

\textbf{Implementation details.}
We implement the \texttt{camroll-agent}'s database with SQLite using two normalized tables: $\mathcal{I}$ for image and their corresponding caption;  and $\mathcal{E}$ for the event.
This two table are linked by a foreign key from \texttt{$\mathcal{I}$.event\_id} to \texttt{$\mathcal{E}$.event\_id}.
On top of this structured store, we build two complementary indices: a BM25 lexical index (SQLite FTS5) for exact matching and verification, and a dense vector index (FAISS) for semantic retrieval under paraphrase or abstraction with ``sentence-transformers/all-MiniLM-L6-v2'' embedding.
This produces a hierarchical fast memory structure consisting of raw images, image-level captions, and event-level summaries, all traceable via stable hashed identifiers (\texttt{img\_<h>} and \texttt{ev\_<h>}).
When building the database, we set the look up window to 3; the tool budget is maximum 25 tools, and the budget for view image is maximum 5 (with maximum of 6 images agent can see at the same time).

\textbf{Baselines.} We benchmark a comprehensive selection of approaches across four families.
(i) \textit{Bare MLLM}: the naive ability of an MLLM with no memory layer, which we feed 4 different inputs: nothing, oracle (gold evidence), all images, and all captions (together with the corresponding timestamps whenever available);
(ii) \textit{RAG-based}: Self-RAG~\cite{selfrag} and HippoRAG-2~\cite{hipporag2}.
(iii) \textit{Memory layer}: SimpleMem~\cite{simplemem}, LightMem~\cite{lightmem}, Mem0~\cite{mem0}, and MemOS~\cite{memos}.
(iv) \textit{AI Agent}: ClaudeCode~\cite{anthropic2025claudecode}, a general-purpose tool-using agent, with a budget of \$0.5 per question.
For a fair comparison, we use GPT-4o-mini~\cite{gpt4} for memory construction and Gemini-2.5-Flash~\cite{gemini} for answering (otherwise specifically required by method). For the all images baseline, we resize each image to a maximum height of 768px to fit Gemini-2.5-Flash's context window and file-upload limit.

\textbf{Metrics.} There are 2 kinds of QA: multi-choice question (MCQ) and freeform. For MCQ, we use accuracy (range 0-100\%); for freeform, we use GPT-4o as judge to compare the predicted answer against the gold answer (range 0-10).
When gold evidence is available, we also report evidence recall, the fraction of gold evidence (images or events) surfaced via tool calls before answering.
We also report input tokens, counted as the cumulative tokens the model consumes (reasoning, input, context, tool calls, and retrieved results) across the entire trace before it emits the final answer.

%% file: sec/6_results.tex
\begin{table}[ht]
\centering
\small
\caption{Quantitative comparison across methods and architectures. Our agent \texttt{camroll-agent} achieves the best results, outperforming all baselines, including bare MLLM with full image captions.
}
\resizebox{\columnwidth}{!}{
\setlength{\tabcolsep}{2.2pt}
\begin{tabular}{@{}>{\raggedright\arraybackslash}p{3.0cm} ll r r rr cc@{}}
\toprule
& & \multicolumn{3}{c}{\textbf{Pre-processing/ Memory Building}} & \multicolumn{2}{c}{\textbf{Multi-choice}} & \multicolumn{2}{c}{\textbf{Free-form}} \\
\cmidrule(lr){3-5} \cmidrule(lr){6-7} \cmidrule(lr){8-9}
\textbf{Method} & \textbf{Base Model} & \textbf{Retrieval Embedding} & \textbf{Build} & \textbf{Tokens$\downarrow$} 
& \textbf{Recall$\uparrow$} & \textbf{Acc$\uparrow$} 
& \textbf{Recall$\uparrow$} & \textbf{Judge$\uparrow$} \\
\midrule

\rowcolor{gray!5}
\multicolumn{9}{l}{\textit{\textbf{Naive LLMs}}} \\
Nothing & Gemini-2.5-Flash & \textit{\textcolor{gray}{no retrieval step}} & 0.0h & $\sim$50 & 0.0 & 30.0 & 0.0 & 0.00 \\
All captions & Gemini-2.5-Flash & \textit{\textcolor{gray}{no retrieval step}} & 1.5h & $\sim$150k & 100.0 & 63.4 & 100.0 & 3.82 \\
All images & Gemini-2.5-Flash & \textit{\textcolor{gray}{no retrieval step}} & 0.0h & $\sim$750k & 100.0 & 76.5 & 100.0 & \underline{5.01} \\
Oracle & Gemini-2.5-Flash & \textit{\textcolor{gray}{no retrieval step}} & 0.0h & $\sim$2.0k & 100.0 & 86.4 & 100.0 & \textbf{6.33} \\

\midrule
\rowcolor{gray!5}
\multicolumn{9}{l}{\textit{\textbf{Retrieval Augmented Generation (RAG)}}} \\
Self-RAG~\cite{selfrag} & LLama-2 & contriever-msmarco & 1.5h & $\sim$2.0k & 25.8 & 46.2 & 19.8 & 2.41 \\
HippoRAG2~\cite{hipporag2} & Gemini-2.5-Flash & NV-Embed-v2 & 1.6h & $\sim$1.0k & 50.1 & 48.5 & 50.1 & 2.58 \\

\midrule
\rowcolor{gray!5}
\multicolumn{9}{l}{\textit{\textbf{Memory Layer}}} \\
SimpleMem~\cite{simplemem} & Gemini-2.5-Flash & Qwen3-Embedding-0.6B & 3.0h & $\sim$0.5k & 57.8 & 44.6 & 58.6 & 1.70 \\
LightMem~\cite{lightmem} & Gemini-2.5-Flash & all-MiniLM-L6-v2 & 1.5h & $\sim$1.0k & 70.3 & 52.7 & 70.2 & 2.44 \\
Mem0~\cite{mem0} & Gemini-2.5-Flash & text-embedding-small-3 & 1.5h & $\sim$1.0k & 75.3 & 53.2 & 75.3 & 2.68 \\
MemOS~\cite{memos} & Gemini-2.5-Flash & BAAI/bge-m3 & 4.0h & $\sim$3.1k & 27.5 & 32.3 & 27.3 & 1.09 \\

\midrule
\rowcolor{gray!5}
\multicolumn{9}{l}{\textit{\textbf{AI Agent}}} \\
ClaudeCode~\cite{anthropic2025claudecode} & Sonnet-4.6 & \textit{\textcolor{gray}{proprietary, unclear trace}} & 0.0h & $\sim$59.0k & -- & \underline{54.0} & -- & \underline{3.77} \\
\rowcolor{green!5}
\texttt{camroll-agent} (ours) & Gemini-2.5-Flash & all-MiniLM-L6-v2 & 1.5h & $\sim$3.2k & 88.5 & \textbf{70.5} & 83.1 & \textbf{4.11} \\

\bottomrule
\end{tabular}
}
\vspace{1mm}

\label{tab:main_results}
\vspace{-4mm}
\end{table}

%

\vspace{-2mm}
\subsection{Comparisons with baselines}

We begin with the naive MLLMs baselines. With no context (\textit{nothing}), performance drops below random on multiple-choice (30\%) and nearly zero on free-form -- unsurprising given the personalized nature of the dataset. Without user-specific information, the model cannot answer meaningfully. At the other extreme, the if the direct gold evidence(s) are given (\textit{oracle}), model performs best (86.4\% multiple-choice), followed by \textit{all images} (5.01) and \textit{all captions} (3.82). This gap exposes two core limitations of the base model: weak long-context reasoning and information loss when compressing images into text. It is worth to note that, these setting in practical would not be possible: \textit{all captions} requires $\sim$150k tokens, while \textit{all images} requires $\sim$750k tokens!


While RAG and memory-layer methods improve over the no-context baseline (40+\% vs. 30\%), they remain well below the \textit{oracle} (86.4\%) and full-context settings (63.4+\%). We hypothesize this is due to limited one-time retrieval: relevant information may be missed or insufficient for complex queries. Additionally, these methods rely on textual representations of images, and thus struggle to capture fine-grained visual details.

Agent-based approaches (ClaudeCode and \texttt{camroll-agent}) surpass all RAG/memory methods. This aligns with their ability to iteratively explore and refine retrieval rather than rely on a single pass. ClaudeCode almost matches \textit{all captions} in free-form performance (3.77 vs. 3.82) while using 2.5 times fewer tokens ($\sim$59k vs. $\sim$150k), showing the benefit of selective exploration. Our \texttt{camroll-agent} goes further, achieving 4.11 with just $\sim$3.2k tokens -- indicating substantially more efficient search and retrieval, thanks to its structured memory and minimal but dedicated set of tool.


\begin{figure}[t]
    \centering
    \begin{minipage}{0.72\linewidth}
        \centering
        \includegraphics[width=\linewidth]{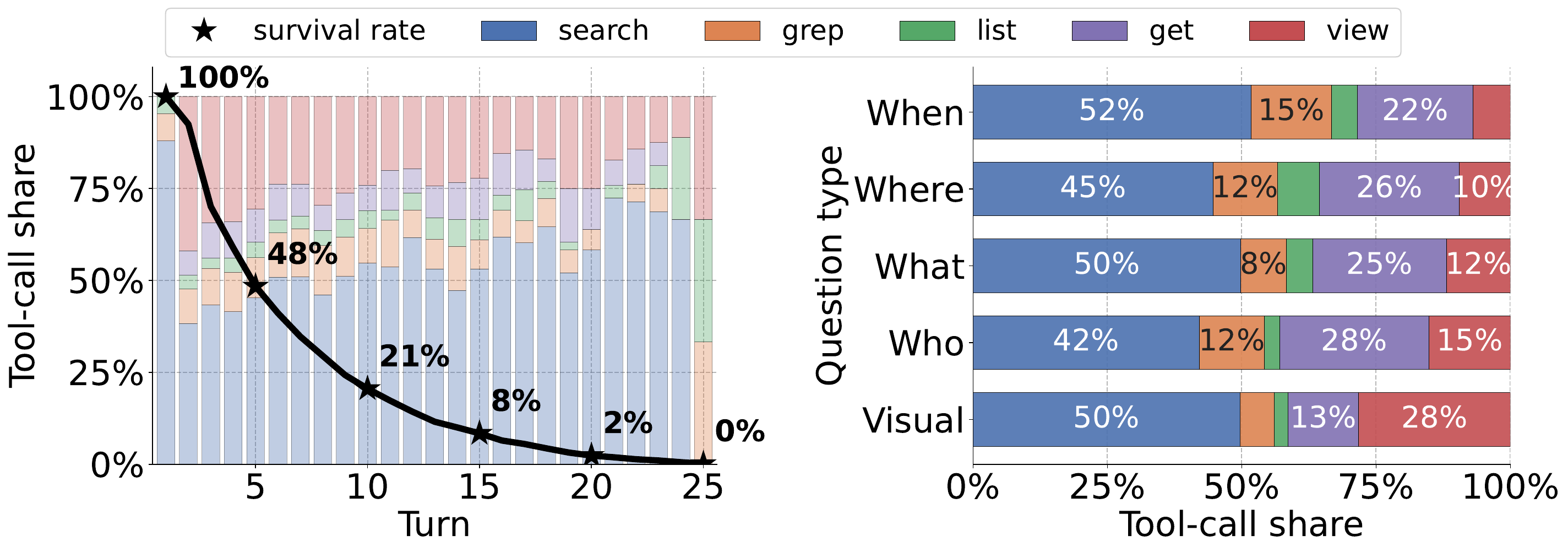}
        \vspace{-3mm}
        \captionof{figure}{Tool-call distributions across turns  and question types.}
        \label{fig:tool_use_per_tool}
    \end{minipage}
    \hfill
    \begin{minipage}{0.27\linewidth}
    \centering
    \renewcommand{\arraystretch}{1.15}
    \resizebox{\linewidth}{!}{
    \setlength{\tabcolsep}{1.5pt}
    \begin{tabular}{lr}
        \toprule
        \textbf{Error category} & \% \\
        \midrule
        A. Wrong evidence                       & 14.7 \\
        B. Right evidence, passed image         & 24.7 \\
        C. Ran out of steps                     & 10.5 \\
        D. Gave up prematurely                  & 21.7 \\
        E. Right evidence, flawed answer         & 17.5 \\
        F. Other                                & 10.9 \\
        \midrule
        \textbf{Total}                                & 100.0 \\
        \bottomrule
    \end{tabular}
    }
    \captionof{table}{Error analysis on incorrect questions.}
    \label{tab:error_analysis}
\end{minipage}
\vspace{-5mm}
\end{figure}

\subsection{Analysis}

\textbf{How agents spend their tool budget.} Figure~\ref{fig:tool_use_per_tool} analyses tool usage over interaction turns. The left panel shows the per-turn distribution of tool calls across all QA episodes, while the black survival curve ($\star$) indicates the fraction of episodes still active at each turn. The first turn is dominated by the coarse retrieval tools---\texttt{search}, \texttt{grep}, and \texttt{list}---showing that agents first perform broad candidate discovery before switching to \texttt{get} and \texttt{view} for detailed inspection and verification. Nearly half of all QA episodes terminate by Turn~5 (48\% still active), suggesting that many questions can be resolved with only a few retrieval rounds. Interestingly, for the small subset of difficult questions that survive into later turns, the proportion of coarse retrieval tools rises again, indicating that agents continue expanding the search space rather than repeatedly inspecting the evidence details. At the final budget-constrained turns, agents tend to rely either on high-yield symbolic retrieval (\texttt{grep}, \texttt{list}) or direct raw-pixel inspection (\texttt{view}) to make a final decision.
\noindent
On the right, \emph{Visual} questions allocate a much larger share to \texttt{view}, \emph{when} questions lean on \texttt{list}, and \emph{what/who} questions are search-heavy. The benchmark therefore exercises genuinely different tool-use skills across question types, not just one retrieval pattern dressed up five ways.


\textbf{When agents fail and why}. To better understand the errors of \texttt{camroll-agent}, we use an LLM judge to inspect the full trajectory (tool calls, retrieved evidence, and final answer) and assign each failure to one of six mutually exclusive categories, as shown in Table~\ref{tab:error_analysis} (see Appendix~\ref{appendix:tab:error} for definitions). Most failures stem from poor agent decisions (A--D) rather than the underlying visual understanding ability (E). A and B show wrong trajectories where the agent either misses relevant images during coarse search or chooses not to open the images. C indicates that the agent is not familiar with the task or user information, thus leading to more complicated situations, and may also suggest potential issues in the memory database. D shows that the agent is overconfident and reaches conclusions too easily. In contrast, only 17.5\% of failures trace back to poor VLM ability. Overall, this suggests that dedicated post-training for memory-agent tasks may be required.

\begin{wrapfigure}{r}{0.48\textwidth}
    \centering
    \vspace{-4mm}
    \includegraphics[width=0.48\textwidth]{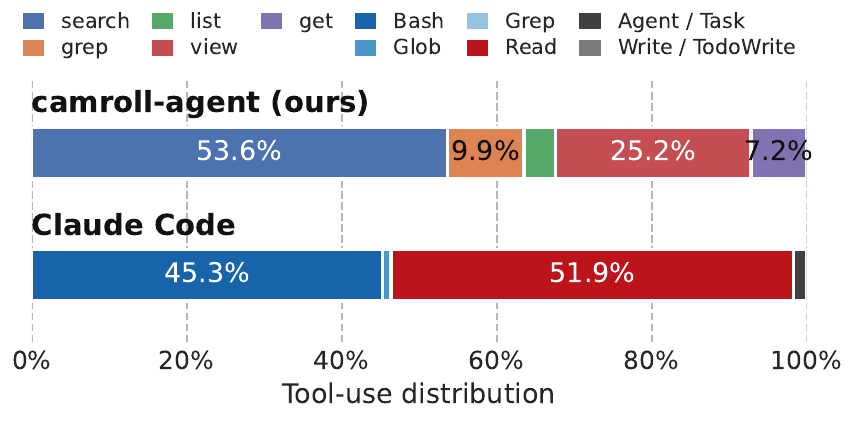}
    \caption{ClaudeCode vs. \texttt{camroll-agent} tool call distributions.}
    \vspace{-4mm}
    \label{fig:vs_claude_code}
\end{wrapfigure}

\textbf{Do we need domain-specific agents?} A generalist coding agent can be repurposed for camera roll setting, but its tool inventory imposes a strong inductive bias toward filesystem traversal and byte-level inspection. As shown in Fig.~\ref{fig:vs_claude_code}, Claude Code lacks a semantic index and therefore alternates between search (e.g., Bash/Glob, 45.3\%) and exhaustive visual inspection (Read, 51.9\%), leading to inefficient investigation of relevant images and yielding high token usage (59.0k). In contrast, our \texttt{camroll-agent} allocates the majority of its budget to a domain-specific semantic retrieval tool (53.6\% \texttt{search}), requiring far fewer image views (25.2\%) (total 3.2k tokens).
This mismatch shows that coding agents can be adapted to new domains, but are inefficient beyond their trained priors and tools. For fundamentally different domains (e.g., visual or continuous), domain-specific tools are not optional but a first-order design choice shaping both behavior and efficiency.

\vspace{-2mm}
\subsection{Ablations}
\label{sec:ablation}
\vspace{-2mm}


\textbf{Memory structure ablation.} As shown in Tab.~\ref{tab:ablation_study}, the full memory design achieves the best performance (i.e., 4.22). Removing structure degrades results consistently: generic captions drop to 4.03 / 4.00 episodic 4.00, no event reduces overall to 4.03, and removing captions causes the largest failure (overall 2.29, episodic 2.04), confirming captions are critical for recall and reasoning.

\begin{table}[h]
\centering
\small

\begin{minipage}{0.52\textwidth}
\centering
\captionof{table}{Comparison of base/build model combinations across proprietary and open-source settings.}
\vspace{1mm}
\setlength{\tabcolsep}{2pt} 
\resizebox{\linewidth}{!}{
\begin{tabular}{llrr}
\toprule
\textbf{Base model} & \textbf{Build model} & \textbf{Input} & \textbf{Judge}$\uparrow$ \\
\midrule
\rowcolor{red!5}
\multicolumn{4}{l}{\textit{\textbf{Proprietary} -- exposure all data}} \\
Gemini-2.5-Flash        & Gemini-2.5-Flash       &   6.3k &  4.12 \\
Gemini-3.1-Preview-Pro  & Gemini-2.5-Flash       &  16.7k &  \textbf{5.80} \\
GPT-4o                  & Gemini-2.5-Flash       &   5.9k &  3.88 \\
GPT-5.2                 & Gemini-2.5-Flash       &   7.6k &  \underline{5.45} \\
\midrule
\rowcolor{red!5}
\multicolumn{4}{l}{\textit{\textbf{Proprietary} -- exposure partial data (build locally)}} \\
Gemini-2.5-Flash        & Qwen3-VL-8B-Instruct   &   6.5k &  3.64 \\
Gemini-3.1-Preview-Pro  & Qwen3-VL-8B-Instruct   &  14.8k &  \textbf{5.30} \\
GPT-4o                  & Qwen3-VL-8B-Instruct   &   5.4k &  3.57 \\
GPT-5.2                 & Qwen3-VL-8B-Instruct   &   7.2k &  \underline{4.99} \\
\midrule
\rowcolor{green!5}
\multicolumn{4}{l}{\textit{\textbf{Open-sourced} -- totally local, private}} \\
Qwen3-VL-8B-Instruct    & Qwen3-VL-8B-Instruct   &   6.0k &  2.05 \\
Qwen3-Coder-30B-A3B     & Qwen3-VL-8B-Instruct   &  13.3k &  \textbf{3.82} \\
GLM-4.7-Flash$^\dagger$ & Qwen3-VL-8B-Instruct   &   8.7k &  \underline{2.60} \\
\bottomrule
\end{tabular}
}
\vspace{1mm}

\label{tab:model_comparison}
\end{minipage}
\hfill
\begin{minipage}{0.47\textwidth}
\centering
\captionof{table}{Ablation study on memory structure and tool usage, reporting semantic and episodic performance, overall score, and efficiency (``J'' denotes LLM-as-Judge score.}
\vspace{1mm}
\setlength{\tabcolsep}{1.0pt}
\resizebox{\linewidth}{!}{
\begin{tabular}{lccccc}
\toprule
\textbf{Setting} & \textbf{Input} $\downarrow$ & \textbf{J-sem}$\uparrow$ & \textbf{J-epi} $\uparrow$ & \textbf{Overall} $\uparrow$ & \textbf{J/inputs} $\uparrow$ \\
\midrule
Final agent             & 2.4k & \textbf{5.90} & \textbf{4.82} & \textbf{4.22} & \textbf{1.24} \\
\midrule
\rowcolor{gray!7}
\multicolumn{6}{l}{\textit{Ablation about memory structure}} \\
Generic caption    & 4.2k & 4.03 & 4.00 & 4.01 & 0.95 \\
No Event         & 3.6k & 4.36 & 3.95 & 4.03 & 1.11 \\
No Caption       & 2.7k & 3.25 & 2.04 & 2.29 & 0.82 \\
\midrule
\rowcolor{gray!7}
\multicolumn{6}{l}{\textit{Ablation about tools}} \\
No Search              & 2.7k & 2.63 & 2.94 & 2.88 & 1.03 \\
No Grep                & 4.0k & 4.25 & 3.90 & 3.97 & 0.97 \\
No List                & 3.7k & 4.28 & 4.04 & 4.09 & 1.08 \\
No Get                 & 4.3k & 4.44 & 3.99 & 4.08 & 0.94 \\
No View                & 3.5k & 3.52 & 3.15 & 3.22 & 0.92 \\
\bottomrule
\end{tabular}
}

\label{tab:ablation_study}
\end{minipage}

\label{tab:combined}
\end{table}

\textbf{Tools ablation.} The full system reaches overall 4.22 (Judge/tokens 1.24, Tab.~\ref{tab:ablation_study}). Removing search causes the largest drop, while removing grep/list/get/view yields smaller but consistent degradations. This shows all tools contribute, with search being the most impactful for performance.

\textbf{MLLMs.} Closed-source models perform best: Gemini-3.1-Preview-Pro achieves the top score (5.80, 5.30; inputs 16.7K / 14.8K across builds), followed by GPT-5.2 (5.45 / 4.99) and Gemini-2.5-Flash (4.12 / 3.64). GPT-4o is lower (3.88 / 3.57). Open-source models lag significantly: Qwen3-VL-8B-Instruct reaches only 2.05, while scaling to Qwen3-Coder-30B-A3B improves to 3.82, still below closed-source systems. While there is a gap, the best open-source model performance is already close to GPT-4o, suggesting a viable alternative for running our agent locally (See Tab.~\ref{tab:model_comparison}).

%% file: sec/7_conclusion.tex
\vspace{-2mm}
\section{Conclusion and Discussion}
\vspace{-2mm}
We introduced \texttt{camroll}, a benchmark for question answering over personal camera rolls, together with \texttt{camroll-agent}, a conversational agent designed for long-horizon personalized visual reasoning. Our results show that hierarchical memory, iterative retrieval, and domain-specific tool use is critical for such task. This work is primarily a benchmark and analysis effort; we do not train a dedicated end-to-end memory agent here. Future work should study learning-based retrieval, joint training, and stronger privacy-preserving personalization.

%% file: sec/appendix.tex
\section{Appendix}

\subsection{Broader Impacts}
This work studies long-horizon reasoning over personal camera rolls, a setting with potential applications in personalized AI assistants, memory support, and multimedia retrieval. At the same time, personal photo collections contain highly sensitive information, including identities, relationships, locations, and daily activities. Systems with persistent multimodal memory therefore raise important privacy and security concerns, including risks of unauthorized retrieval, profiling, or memorization of personal content. Future deployments should prioritize user consent, controllable memory management, secure storage, and privacy-preserving mechanisms. We hope this work encourages further research on safe and transparent personalized multimodal memory systems.

\subsection{Data Statistic}
\label{appendix:sec:dataset}

\textbf{Demographics and coverage.} Fig.~\ref{fig:combined_demographic} summarizes the geographic and temporal footprint of the Camroll dataset. Together, the two subsets span 24 years of personal photo-taking (2002--2026) across 05 continents and roughly 25 countries). The right panel (Fig.~\ref{fig:combined_demographic}b) shows that the two subsets capture two distinct patterns: YFCC peaks in 2006--2010 with the global popularity of dedicated digital cameras, while the in-house subset accelerates sharply from 2023 onward, reflecting the dense, low-curation regime of contemporary smartphones. Per active user, the smartphone era is roughly 1.6$\times$ denser than the digital-camera era: mobile users accumulate on average $\sim$17 photos/month, versus $\sim$11/month for YFCC100M users. This is clearly that while two subsets of comparable absolute size (15,658 vs.\ 15,607 images) but very different temporal density profiles.

\textbf{Question analysis.} We analyze the questions set to understand its linguistic structure and grounding requirements. Episodic questions are roughly 2$\times$ longer than semantic ones (15.9 vs. 7.3 words on average), reflecting the additional contextual information needed to specify time, place, or events. Beyond length, we observe clear differences in question formulation: semantic questions are dominated by \emph{what}-style queries (57\%), whereas episodic questions are more diverse and heavily shaped by temporal and prepositional phrasing (e.g., ``on'', ``in'', ``after'', ``during'').
About 46.2\% of questions can be answered from a single image, while 32.2\% require reasoning across multiple images and 20.0\% require whole-roll context. This shows that more than half of questions goes beyond single-image VQA and requires cross-image or sequence-level reasoning. In addition, 23.8\% of questions involve fine-grained perceptual understanding (e.g., counting, OCR, or attribute-level details). Finally, Camroll is strongly first-person centered (88.4\% explicit ``I/my/me'' usage) and temporally grounded (62.4\% contain explicit time or event references), reinforcing its nature as a personal, longitudinal memory benchmark rather than standard visual question answering.

\textbf{Answers analysis.}
Gold answers are typically short but rarely single-word: the median answer length is 2 tokens, and 72.9\% are multi-word phrases (mean 2.86, max 15). This reflects that questions about personal life often require precise answers (e.g., a specific outfit, a place name, or a duration), rather than a single object label or overly long descriptions (e.g., image captioning). Episodic answers are slightly longer than semantic ones (mean 2.95 vs.\ 2.51 tokens), consistent with the need to disambiguate among similar past events. Distractors are written by the same annotator with knowledge of the user's roll, and 89.7\% are length-matched to the gold answer to within two tokens, so the format does not leak the correct option through surface form.
Most importantly, gold answers are \emph{personal}: Of the 2,084 distinct content tokens (length $\geq$ 4) that appear in gold answers, 66.9\% appear in only one user's answers, and 88.2\% of unique answer bigrams appear in only a single user's roll. Of the 1{,}875 distinct full answer strings, only 9.8\% are reused across two or more users., and the most frequently repeated answers are exactly those with weak personalization signal (\textit{white}, \textit{student}, \textit{yellow}, \textit{red}). This indicates that solving CamRoll requires retrieving content from the target user's own album rather than relying on common visual concepts shared across users. 


\begin{table}[ht]
\centering

\caption{Evidence coverage across dataset subsets and memory types.}
\label{tab:evidence_coverage}
\vspace{1mm}

\begin{tabular}{l l c c}
\toprule
\textbf{Subset} & \textbf{Type} & \textbf{\% with evidence} & \textbf{n} \\
\midrule
In-house & semantic & 0\% & 0 / 300 \\
In-house & episodic & 97.4\% & 1,169 / 1,200 \\
YFCC & semantic & 0\% & 0 / 200 \\
YFCC & episodic & 98.9\% & 791 / 800 \\
Combined & all & 78.4\% & 1,960 / 2,500 \\
Combined & episodic only & 98.0\% & 1,960 / 2,000 \\
\bottomrule
\end{tabular}

\end{table}

\begin{table}[t]
\centering
\caption{Distribution of licenses in 20 users of YFCC-100M}
\vspace{1mm}
\small
\begin{tabular}{lrr}
\toprule
\textbf{License} & \textbf{Count} & \textbf{\%} \\
\midrule
CC BY-NC-SA 2.0 & 5,438 & 34.1 \\
CC BY 2.0       & 4,945 & 31.0 \\
CC BY-NC 2.0    & 4,028 & 25.3 \\
CC BY-SA 2.0    & 1,516 & 9.5 \\
\midrule
\textbf{Total}  & \textbf{15,927} & \textbf{100.0} \\
\bottomrule
\end{tabular}

\label{tab:yfcc_license_distribution}
\end{table}


\begin{table}[ht]
\centering
\small
\caption{Composition of \textsc{Camroll}. The two subsets are complementary: the in-house subset captures contemporary smartphone behavior at full resolution with rich participant-authored event labels, while YFCC contributes longer per-user spans, real EXIF/GPS metadata, and a publicly redistributable license at lower resolution. $^*$Encoded in the filename (\texttt{YYYY-MM-DD HHMMSS.jpg}); $^\dagger$encoded in the YFCC100M \texttt{datetaken} metadata field; $^\ddagger$YFCC also contains a smaller fraction of early-smartphone captures (e.g., iPhone 4).}
\vspace{1mm}
\begin{tabular}{l rrr}
\toprule
\textbf{Property} & \textbf{In-house} & \textbf{YFCC} & \textbf{Total} \\
\midrule
\multicolumn{4}{l}{\textit{Users \& images}} \\
Number of users                       & 30           & 20           & 50           \\
Total images                          & 15{,}869     & 15{,}607     & 31{,}476     \\
Images per user (mean / median)       & 529 / 533    & 780 / 808    & 630 / 558    \\
Images per user (min / max)           & 225 / 707    & 505 / 979    & 225 / 979    \\
\midrule
\multicolumn{4}{l}{\textit{Temporal coverage}} \\
Capture date range                    & 2019\,--\,2026 & 2002\,--\,2014 & 2002\,--\,2026 \\
Per-user span, years (mean)           & 2.7          & 6.6          & 4.3          \\
Per-user span, years (min / max)      & 1.6 / 3.8    & 3.0 / 9.4    & 1.6 / 9.4    \\
Images with second-precision timestamp & 98\%$^*$    & 100\%$^\dagger$ & 99\%       \\
\midrule
\multicolumn{4}{l}{\textit{Image properties}} \\
Resolution, mean (W $\times$ H)       & 3188 $\times$ 3500 & 592 $\times$ 514 & --   \\
Resolution, median (W $\times$ H)     & 3072 $\times$ 4000 & 500 $\times$ 480 & --   \\
Megapixels (mean / median)            & 11.7 / 12.2  & 0.43 / 0.19  & --           \\
Orientation (landscape / portrait / square) & 33\% / 64\% / 3\% & 60\% / 23\% / 17\% & 46\% / 44\% / 10\% \\
File size, median                     & 3.0 MB       & 61 KB        & --           \\
Total dataset size                    & 64 GB        & 2.3 GB       & 66 GB        \\
\midrule
\multicolumn{4}{l}{\textit{Capture metadata}} \\
Capture device                        & smartphone   & digital camera$^\ddagger$ & mixed \\
Geo-tagged images                     & --           & 75\%         & 37\%         \\
Profile photo                         & \checkmark   & \checkmark   & \checkmark   \\
Profile biography (text)              & --           & 13 / 20 users & --          \\
\midrule
\multicolumn{4}{l}{\textit{Annotations}} \\
User-labeled events                   & 841 (28/user)& --           & 841          \\
Avg.\ event size (\# images)          & 9.7          & --           & --           \\
QA pairs (semantic + episodic)        & 300 + 1{,}200 & 200 + 800   & 500 + 2{,}000 \\
QA pairs, total                       & 1{,}500      & 1{,}000      & \textbf{2{,}500} \\
Episodic Qs grounded to image(s)      & 97\%         & 99\%         & 98\%         \\
\bottomrule
\end{tabular}

\label{tab:camroll-stats}
\end{table}

\begin{table}[t]
\centering
\small
\caption{Question-type schema. Each question is assigned exactly one label
by an LLM classifier (\textsc{gemini-2.5-flash}), based on the shape of
the gold answer rather than on the question's surface form. Tie-breaking
priority: \textsc{Visual} $>$ \textsc{When} $>$ \textsc{Where} $>$
\textsc{Who} $>$ \textsc{What}.}
\label{tab:qtype-schema}
\begin{tabular}{@{}lp{0.22\linewidth}p{0.38\linewidth}r@{}}
\toprule
\textbf{Label} & \textbf{Answer is\ldots} & \textbf{Example question} & \textbf{$n$} \\
\midrule
\textit{What}   & an object or action       & ``What did I eat for breakfast?''                                       & 611 \\
\textit{Where}  & a place, venue, or location & ``Where did I eat dinner the day before going to the museum?''        & 173 \\
\textit{When}   & a date, duration, or temporal order & ``When did I last see my grandfather?''                       & 123 \\
\textit{Who}    & a person or group         & ``Who came to my birthday party in 2024?''                              & 75  \\
\textit{Visual} & a visual attribute or exact count from a photo (color, count, written text, breed, fine-grained appearance) & ``How many balloons were in the photo?'' & 518 \\
\midrule
\multicolumn{3}{r}{\textbf{Total}} & 1{,}500 \\
\bottomrule
\end{tabular}
\end{table}

\begin{table}[t]
\centering
\small
\caption{Condition-type schema. Conditions are the constraints in the
question that scope the search (which photo / event / email to look at),
separately from the question type (what attribute to extract). A single
question may carry zero, one, or several conditions; counts therefore
sum to more than 1{,}500. The condition vocabulary is intentionally
distinct from the question-type vocabulary so the two slots are not
conflated. }
\label{tab:cond-schema}
\begin{tabular}{@{}lp{0.20\linewidth}p{0.40\linewidth}r@{}}
\toprule
\textbf{Label} & \textbf{Triggered by\ldots} & \textbf{Example phrase in the question} & \textbf{$n$} \\
\midrule
\textit{Situation} & an episode, activity, or object that scopes the search & ``the day I visited the museum'', ``during my road trip'', ``at my friend's wedding'' & 1{,}182 \\
\textit{Location}  & a place, venue, or location reference                  & ``in Beijing'', ``at the zoo'', ``in the kitchen''                                    & 621   \\
\textit{Time}      & a date, year, time of day, or temporal anchor          & ``in 2024'', ``the day after'', ``on Christmas''                                       & 778   \\
\textit{Person}    & a specific person whose presence scopes the search     & ``with my mom'', ``when I was with Lin''                                              & 423   \\
\midrule
\textit{(none)} & \multicolumn{2}{l}{question imposes no scoping constraint, e.g. ``What did I eat for breakfast?''} & 235 \\
\bottomrule
\end{tabular}
\end{table}

\paragraph{Error categorization.}
We classify each \emph{incorrectly}-answered question (LLM-judge score
\(=0/10\)) into one of six mutually-exclusive categories, evaluated in the
order listed.\footnote{Categories are evaluated top-to-bottom; the first
matching rule wins, so each question lands in exactly one bucket.} Let
\(s\) denote the number of agent actions in the trace, \(v\) the number of
\texttt{view\_image} calls, and \(\rho\in[0,1]\) the
\textsc{recall\_img\_or\_event} signal: the fraction of ground-truth
evidence images that appeared (by stem or via a containing event name) in
\emph{any} tool result of the trace. Let
\(\mathcal{G}\) be the set of ground-truth evidence image stems and
\(\mathcal{V}\) the set of stems the agent actually opened with
\texttt{view\_image}.

\begin{description}
  \item[\textbf{(c) Ran out of steps / budget.}]
    The agent exhausted its action budget: either
    \(\texttt{stopped\_reason}=\texttt{max\_steps}\), or it used at least
    \(20\) actions, or it hit the per-trace \texttt{view\_image} cap of
    \(5\) calls. These traces ended because of a hard limit, not because
    the agent decided it was done.
  \item[\textbf{(d) Gave up prematurely.}]
    The agent voluntarily stopped (\(\texttt{stopped\_reason}=\texttt{ok}\))
    after at most \(\texttt{PREMATURE\_STEPS}=2\) tool calls. The agent
    answered with very little exploration.
  \item[\textbf{(a) Wrong evidence.}]
    Ground-truth evidence exists but the trace failed to retrieve it
    (\(\rho<1\)). The retrieval pipeline did not surface all of the right
    images / events.
  \item[\textbf{(b\textsubscript{1}) Right evidence, looked, still wrong.}]
    All gold evidence was retrieved (\(\rho=1\)) \emph{and} the agent
    explicitly opened at least one gold image with \texttt{view\_image}
    (\(\mathcal{G}\cap\mathcal{V}\neq\emptyset\)), yet still produced a
    wrong answer. This is a genuine perception-detail or reasoning
    failure on inspected content.
  \item[\textbf{(b\textsubscript{2}) Right evidence, never looked.}]
    All gold evidence was retrieved (\(\rho=1\)) but the agent never
    invoked \texttt{view\_image} on any of the gold images
    (\(\mathcal{G}\cap\mathcal{V}=\emptyset\)). The agent answered
    over-confidently on a search-result snippet without inspecting the
    image itself.
  \item[\textbf{(e) Other.}]
    The question carries no ground-truth evidence list, or the evidence
    signal is unavailable, so the (a)/(b\textsubscript{1})/(b\textsubscript{2})
    distinction does not apply (e.g., semantic questins)
\end{description}

\clearpage
\subsection{Prompts}
\label{appendix:sec:prompt}
\begin{tcolorbox}[
    colback=promptbg,
    colframe=promptborder,
    coltitle=black,
    boxrule=0.5pt,
    arc=3pt,
    left=3pt,
    right=3pt,
    top=4pt,
    bottom=4pt,
    title=Image Captioning and Event Segmentation Prompt,
    fonttitle=\bfseries
]
You are maintaining long-term structured memory for one user's personal photo library.

You will see two images:
1. The user's profile photo.
2. The current image that must be processed.

Important:
- Use the first image only as identity/reference image. This image will tell you how the user looks like.
- You should write the caption and event decision from the perspective of the user in the first image.
- The second image as the only source for ``image-caption'', ``operation'', and event reasoning.
- Never describe the first image in ``image-caption''.
- The album is processed strictly in chronological order from oldest to newest.
- Only update the most recent event row if the current image clearly belongs to it.

Event definition:
- An event should be an episodic memory unit, such as a trip, outing, meal, hangout, celebration, class activity, or other coherent real-world episode.
- An event can span multiple consecutive photos with different dates, locations, subjects, poses, or close-up details, as long as they still belong to the same broader episode (e.g., a road trip, hangout with friends, selfies, etc.).
- Event names should summarize the broader episode, not the most eye-catching object in a single frame.

Album description:
\texttt{\detokenize{\{library-description\}}}

Current image metadata:
- date: \texttt{\detokenize{\{current-date\}}}

Recent image table rows (up to the last k images):
\begin{verbatim}
json.dumps(recent-payload, indent=2, ensure_ascii=False)
\end{verbatim}

Latest event summary:
\begin{verbatim}
json.dumps(latest_event_payload, indent=2, ensure_ascii=False)
\end{verbatim}

Tasks:
1. Write a detailed personalized caption for image 2 (the current album image) as if the person in image 1 is describing their own photo in first person.

2. Choose exactly one operation for the event table:
   - ADD: create a new event row
   - UPDATE: update the latest event row
   - NO OP: do not modify the event table

3. If you choose ADD or UPDATE, return a full event row with fields:
   - event name
   - description
   - date
   - images

Rules:
- The image caption must always be present.
- Use first-person wording when natural.
- The image caption must describe image 2 only, not the profile/reference image.
- Be careful about person identity; the person in the first image is the user, and the person in the second image might be different.
- If identity is unknown, use neutral terms like "a man", "a woman", or inferred relations like "my friend".
- Make the image caption specific and detailed.
- Mention visible content such as setting, people, objects, activity, and atmosphere.
- Write as a personal memory grounded only in image 2.
- Do not hallucinate precise facts not supported by the image.

Event update rules:
- For ADD or UPDATE, include current image path in images.
- Prefer UPDATE if the image is part of the same ongoing episode.
- Prefer ADD only when a clear new event boundary exists.
- Keep event descriptions under 300 words and concise.
- Event names should reflect broad activities (e.g. "Trip to Chengdu", "Coffee Shop Hangout").
- Prefer NO-OP if nothing meaningful changes.

Output format (VALID JSON ONLY, no markdown):

\begin{verbatim}
{
  "operation": "ADD" | "UPDATE" | "NO_OP",
  "image_caption": "string",
  "event": {
    "event_name": "string",
    "description": "string",
    "date": "string",
    "images": ["full image path", "..."]
  } | null
}
\end{verbatim}
\end{tcolorbox}